\documentclass[a4paper]{article}

\usepackage{INTERSPEECH_v2}
\usepackage{gensymb}
\usepackage{siunitx}
\RequirePackage{graphicx,subfigure}

\title{End-to-End Audiovisual Fusion with LSTMs \thanks{Accepted to Auditory-Visual Speech Processing Conference 2017}} 
\name{Stavros Petridis$^1$, Yujiang Wang$^1$, Zuwei Li$^1$, Maja Pantic$^{1,2}$}
\address{
  $^1$Dept. Computing, Imperial College London\\
  $^2$EEMCS, University of Twente}
\email{stavros.petridis04@imperial.ac.uk, m.pantic@imperial.ac.uk}

\begin{document}

\maketitle
\begin{abstract} 
Several end-to-end deep learning approaches have been recently presented which simultaneously extract visual features from the input images and perform visual speech classification. However, research on jointly extracting audio and visual features and performing classification is very limited. In this work, we present an end-to-end audiovisual model based on Bidirectional Long Short-Term Memory (BLSTM) networks. To the best of our knowledge, this is the first audiovisual fusion model which simultaneously learns to extract features directly from the pixels and spectrograms and perform classification of speech and nonlinguistic vocalisations.  The model consists of multiple identical streams, one for each modality, which extract features directly from mouth regions and spectrograms. The temporal dynamics in each stream/modality are modeled by a BLSTM and the fusion of multiple streams/modalities takes place via another BLSTM. An absolute improvement of 1.9\% in the mean F1 of 4 nonlingusitic vocalisations over audio-only classification is reported on the AVIC database. At the same time, the proposed end-to-end audiovisual fusion system improves the state-of-the-art performance on the AVIC database leading to a 9.7\% absolute increase in the mean F1 measure.  We also perform audiovisual speech recognition experiments on the OuluVS2 database using different views of the mouth, frontal to profile. The proposed audiovisual system significantly outperforms the audio-only model for all views when the acoustic noise is high.

\end{abstract}

\begin{figure}[th]
  \centering
\includegraphics[width=\linewidth]{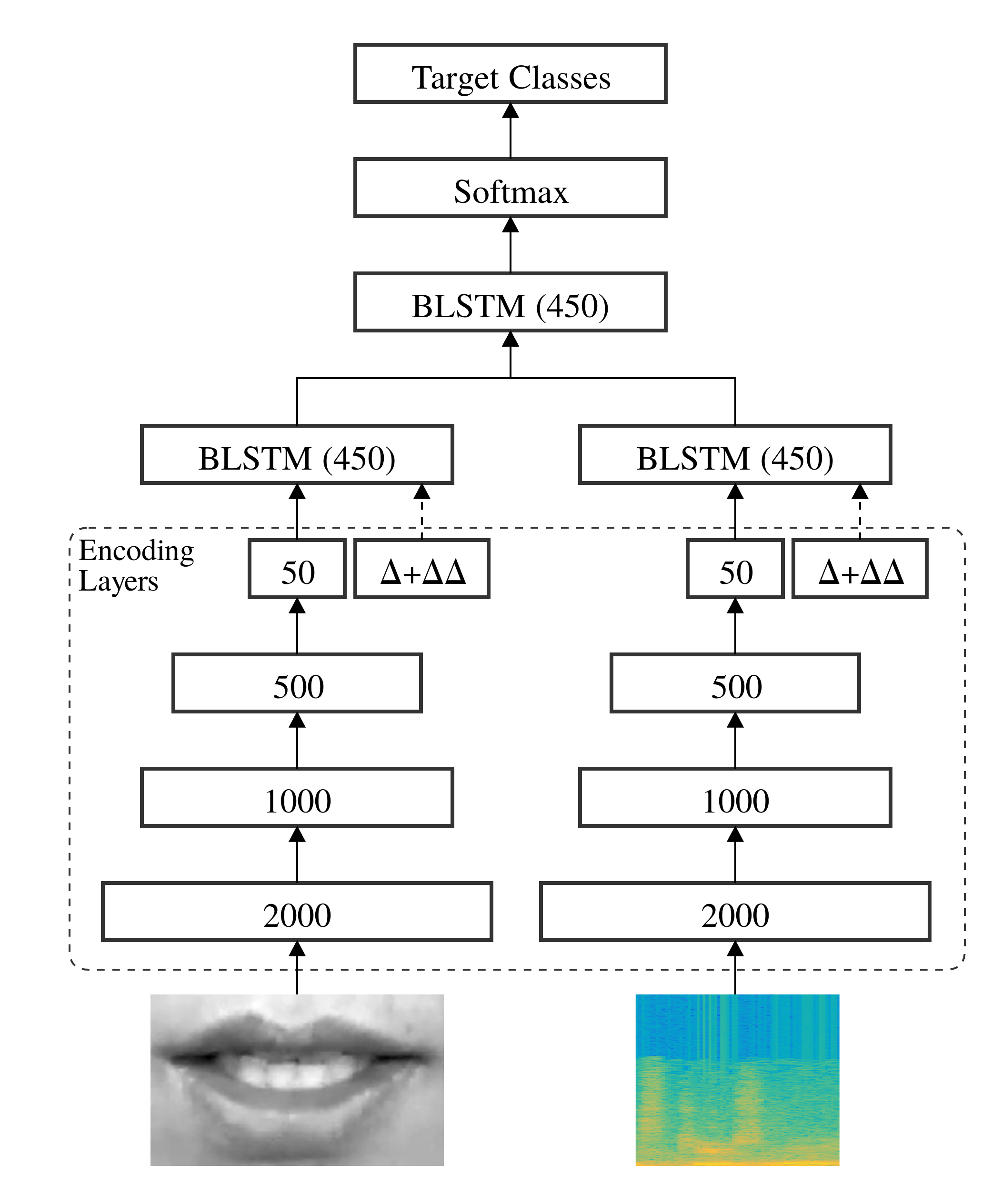}
  \caption{Overview of the end-to-end audiovisual system. One stream per modality is used for feature extraction directly from the raw images and spectrograms. Each stream consists of an encoder which compresses the high dimensional input image to a low dimensional representation. The $\Delta$ and $\Delta\Delta$ features are also computed  and appended to the bottleneck layer. The encoding layers in each stream are followed by a BLSTM which models the temporal dynamics. A BLSTM is used to fuse the information from all streams and provides a label for each input frame. }
\label{fig:system}
\end{figure}

\section{Introduction  }

Audiovisual fusion approaches have been successfully applied to various problems like speech recognition \cite{Dupont2000, Potamianos2003}, emotion recognition \cite{ZengPantic2008,gunes2013categorical}, laughter recognition \cite{petridis2011TMM} and biometric applications \cite{aleksic2006audio}. 
The addition of the visual modality is particularly useful in noisy environments where the performance of audio-only classifiers is degraded. As a consequence, 
the visual information, which is not affected by acoustic noise, can significantly improve the performance of audio-only classifiers in noisy environments. 

Recently, several deep learning approaches for audiovisual fusion have been presented. The vast majority of them follow a two step approach where features are first extracted from the audio and visual modalities and then are fed to a classifier.  Ngiam et al. \cite{ngiam2011multimodal} applied principal component analysis (PCA) to the mouth region of interest (ROI) and spectrograms and trained a deep autoencoder to extract bottleneck features. The features from the entire utterance were fed to a support vector machine (SVM) ignoring the temporal dynamics of the speech. Hu et al. \cite{hu2016temporal} used a similar approach where PCA was applied to mouth ROIs and spectrograms and a recurrent temporal multimodal restricted Boltzmann machine was trained to extract features which are fed to an SVM. Ninomiya et al. \cite{ninomiya2015integration} applied PCA to the mouth ROIs and concatenated Mel-Frequency Cepstral Coefficients (MFCCs) and trained a deep autoencoder to extract bottleneck features which were fed to a Hidden Markov Model (HMM) in order to take into account the temporal dynamics.  Mroueh et al. \cite{mroueh2015} used concatenated MFCCs together with scattering coefficients extracted from the mouth ROI in order to train a deep network with a bilinear softmax layer. Takashima et al. \cite{takashima2016audio} used a convolutional neural network to extract bottleneck features from lip images and Mel-maps which were fed to an HMM. It is clear that none of the above works follows an end-to-end architecture.

Few works have been presented very recently which follow an end-to-end approach for visual speech recognition (lipreading). Wand et al. \cite{wand2016lipreading} used a fully connected layer followed by two LSTM layers to perform lipreading directly from raw mouth ROIs. Petridis et al. \cite{petridis2017deepVisualSpeech} used a deep autoencoder together with an LSTM for end-to-end lipreading from raw pixels. Assael et al. \cite{assael2016lipnet} used a CNN with gated recurrent units for end-to-end sentence-level lipreading. 

To the best of our knowledge, the only work which performs end-to-end training for audiovisual speech recognition is \cite{chung2016lipSentences}. An attention mechanism is applied to both the mouth ROIs and MFCCs and the model is trained end-to-end. However, the system does not use the raw audio signal or spectrogram but relies on  MFCC features.

In this paper, we extend our previous work \cite{petridis2017deepVisualSpeech} and present an end-to-end audiovisual fusion model for speech recognition and nonlinguistic vocalisation classification which jointly learns to extract  audio/visual features directly from raw inputs and perform classification (Fig. \ref{fig:system}). To the best of our knowledge, this is the first end-to-end model which performs audiovisual fusion from raw mouth ROIs and spectrograms. The proposed model consists of multiple identical streams, one per modality,  which extract features directly from the raw images and spectrograms. Each stream consists of an encoder which compresses the high dimensional input  to a low dimensional representation. The encoding layers in each stream are followed by a BLSTM which models the temporal dynamics. Finally,  the information of the different streams/modalities is fused  via a BLSTM which also provides a label for each input frame. We perform classification of nonlinguistic vocalisations on AVIC database achieving state-of-the-art performance for audiovisual fusion, with an absolute increase in the mean F1 measure by 9.7\%. The proposed system also results in an absolute increase of 1.9\% in the mean F1 measure compared to the audio-only model. In addition, we also perform experiments on audiovisual speech recognition using different lip views, from frontal to profile, on OuluVS2. The end-to-end audiovisual fusion outperforms the audio-only model when the noise level is high and results in the same performance when clean audio is used.

\section{Databases}
\label{sec:dbs}

The databases used in this study are the OuluVS2 \cite{Anina2015} and AVIC \cite{schuller2009being}. The OuluVS2 contains 52 speakers saying 10 utterances, 3 times each, so in total there are 156 examples per utterance. The utterances are the following: ``Excuse me", ``Goodbye", ``Hello", ``How are you", ``Nice to meet you", ``See you", ``I am sorry", ``Thank you", ``Have a good time", ``You are welcome".  The mouth ROIs are provided and they are downscaled as shown in Table \ref{tab:mouthROIsize} in order to keep the aspect ratio of the original videos constant. Video is recorded at 30 frames per second (fps) and audio at 48 \si{\kilo\hertz} The unique feature of OuluVS2 is that it provides multiple lip views. To the best of our knowledge it is the only publicly available database with 5 lip views between 0\degree and 90\degree. The LiLir dataset \cite{Lan2012ViewInd} also contains five views but it is not publicly available at the moment, and the TCD-Timit database \cite{harte2015tcd} contains only two views, frontal and 30\degree. 

The AVIC corpus is an audiovisual dataset containing scenario-based dyadic interactions. A subject is interacting with an experimenter who plays the
role of a product presenter and leads the subject through a commercial
presentation. The subjects role is to listen to the presentation and interact with the experimenter depending on his/her interest on the product. 

\begin{figure*}[t]

\begin{minipage}[t]{0.5\textwidth}
  \centering
  \subfigure[6532]{\includegraphics[width = 0.24\columnwidth] {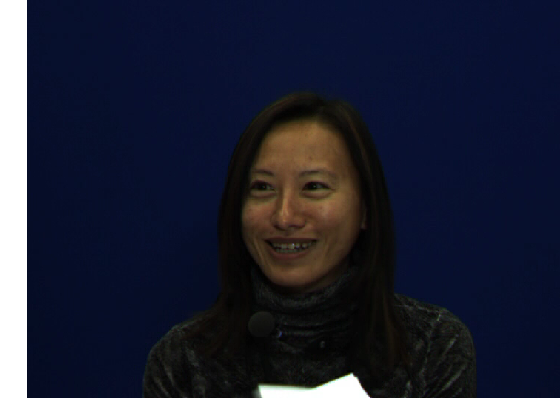}}  
  \subfigure[6539]{\includegraphics[width = 0.24\columnwidth] {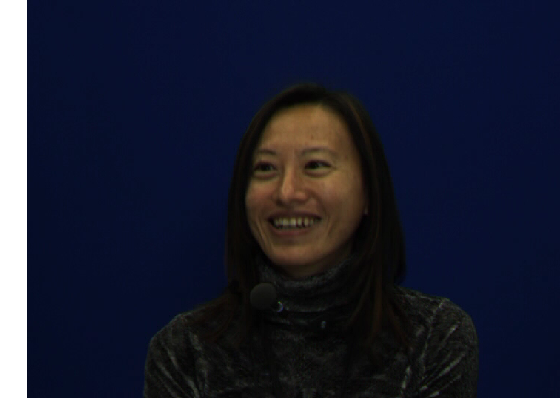}}   
  \subfigure[6555]{\includegraphics[width = 0.24\columnwidth] {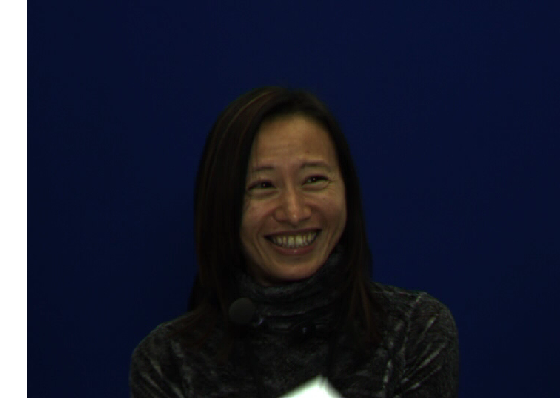}}   
  \subfigure[6569]{\includegraphics[width = 0.24\columnwidth] {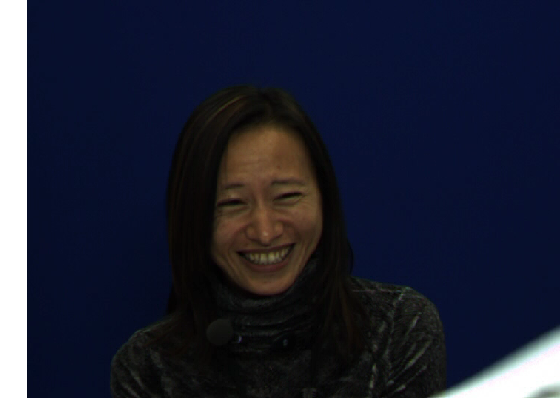}} 

  \caption{Example of laughter from the AVIC corpus, Subject VP4, frames 6532 to 6569.}
  \label{fig:laughterExample_AVIC}
\end{minipage}
\begin{minipage}[t]{0.5\textwidth}
  \centering
  \subfigure[15476]{\includegraphics[width = 0.24\columnwidth] {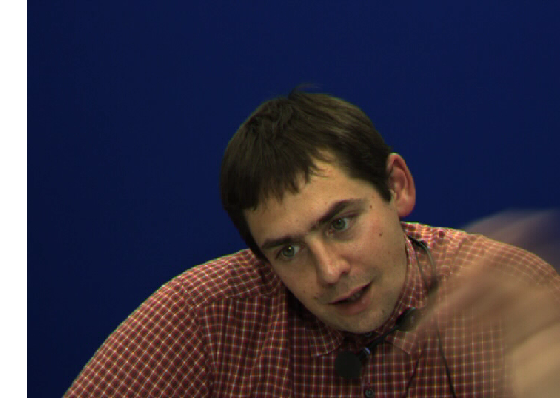}}                
  \subfigure[15479]{\includegraphics[width = 0.24\columnwidth] {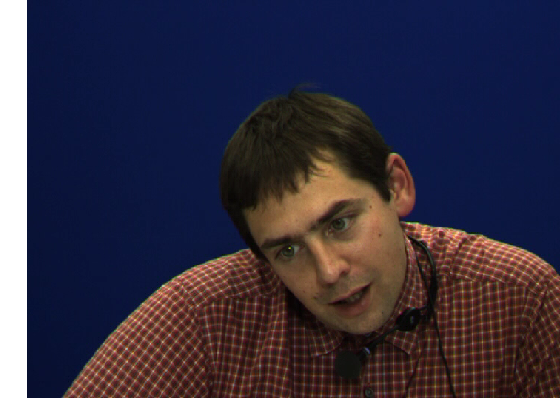}}     
  \subfigure[15489]{\includegraphics[width = 0.24\columnwidth] {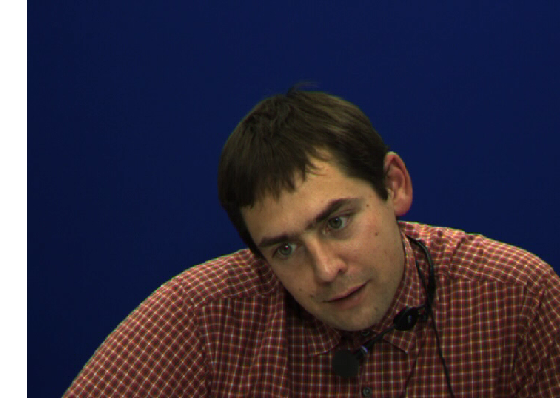}}   
  \subfigure[15497]{\includegraphics[width = 0.24\columnwidth] {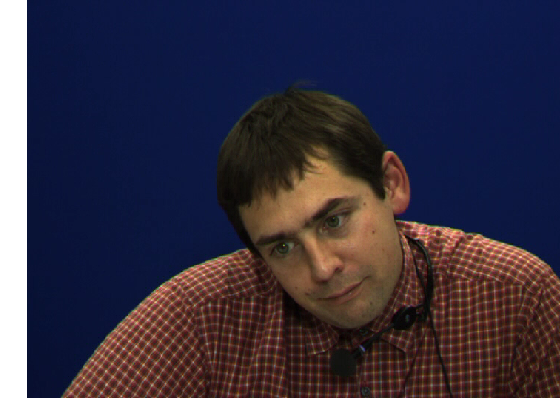}} 

  \caption{Example of hesitation from the AVIC corpus, Subject VP8, frames 15476 to 15497.}
  \label{fig:hesitationExample_AVIC}
\end{minipage}

\end{figure*}

Annotations for laughter, hesitation, consent and other human noises, which are grouped into one class called garbage, are provided with the database and those are used in this study. In total 21 subjects were recorded, 11 males and 10 females with most subjects being non-native speakers. Similarly to previous works \cite{schuller2009being, eyben2011AV-vocalOutb,petridis2016Pred} vocalisations that were very short ($\leq$ 120 \si{\milli\second}) were excluded. In total, 247, 1136, 308 and 582 examples for the laughter, hesitation, consent and garbage class, respectively, were used. Examples of laughter and hesitation 
are shown in Fig. \ref{fig:laughterExample_AVIC} and \ref{fig:hesitationExample_AVIC}, respectively. 

A video camera was used to record the subject's reaction, positioned in front of him/her, at 25 fps. The audio signal was recorded by a lapel microphone at 44.1 \si{\kilo\hertz}.

AVIC does not provide mouth ROIs so sixty eight points were tracked on the face using the tracker proposed in \cite{Kazemi_2014_CVPR}. The faces were first aligned using a neutral reference frame in order to normalise them for rotation and size differences. This is done using an affine transform using five stable points, two eyes corners in each eye and the tip of the nose. Then the center of the mouth is located based on the tracked points and a bounding box with size 85 by 129 is used to extract the mouth ROI. Finally, the mouth ROIs are downscaled to 30 by 45.

\begin{table}[tb]
\renewcommand{\arraystretch}{1.1}
\caption{Size of mouth ROIs in pixels for each view in the OuluVS2 database.}
\label{tab:mouthROIsize}
\centering
\begin{tabular}{cccccc}
\toprule  Views & \multicolumn{1}{c}{0\degree}  & \multicolumn{1}{c}{30\degree} & \multicolumn{1}{c}{45\degree} & \multicolumn{1}{c}{60\degree} & \multicolumn{1}{c}{90\degree}  \\

\midrule Height/Width & 29/50 & 29/44 & 29/43 & 35/44 & 44/30 \\

\bottomrule

\end{tabular} 

\end{table}

\section{End-to-end Audiovisual Fusion}
The proposed deep learning system for audiovisual fusion is shown  in Fig. \ref{fig:system}. It consists of two identical streams which extract features directly from the raw input images and the spectrograms\footnote{Spectrogram frame are computed over a 40 \si{\milli\second} windows with 30 \si{\milli\second} overlap. }, respectively. Each stream consists of two parts: an encoder and a BLSTM. The encoder follows a bottleneck architecture in order to compress the high dimensional input image to a low dimensional representation at the bottleneck layer.  The same architecture as in \cite{hinton2006reducing} is used, with 3 hidden layers of sizes 2000, 1000 and 500, respectively, followed by a linear bottleneck layer. The rectified linear unit is used as the activation function for the hidden layers. The $\Delta$ (first derivatives) and $\Delta\Delta$ (second derivatives) \cite{young2002htk} features are also computed, based on the bottleneck features, and they are appended to the bottleneck layer. In this way, during training we force the encoding layers to learn compact representations which are discriminative for the task at hand but also produce discriminative $\Delta$ and $\Delta\Delta$ features. This is in contrast to the traditional approaches which pre-compute the $\Delta$ and $\Delta\Delta$ features at the input level and as a consequence there is no control over their discriminative power. 

The second part is a BLSTM layer added on top of the encoding layers in order to model the temporal dynamics of the features in each stream. The BLSTM outputs of each stream are concatenated and fed to another BLSTM in order to fuse the information from all streams. The output layer is a softmax layer which provides a label for each input frame. The majority label over each utterance is used in order to label the entire utterance. The entire system is trained end-to-end which enables the joint learning of features and classifier. In other words, the encoding layers learn to extract features from raw images and spectrograms which are useful for classification using BLSTMs.

\section{EXPERIMENTAL SETUP}

\subsection{Evaluation Protocol}

We first partition the data into training, validation and test sets. The same protocol as in \cite{petridis2016Pred} is used for the AVIC dataset where the first 7 subjects are used for testing, the next 7 for training and the last 7 for validation.

The protocol suggested in \cite{saitoh2016concatenated} is used for the OuluVS2 dataset where 40 subjects are used for training and validation and 12 for testing. We randomly divided the 40 subjects into 35 and 5 subjects for training and validation purposes, respectively.  This means that there are 1050 training utterances, 150 validation utterances and 360 test utterances.

\subsection{Preprocessing}

Since all the experiments are subject independent we first need to reduce the impact of subject dependent characteristics. This is done by subtracting the mean image, computed over the entire utterance, from each frame.

 As mentioned in section \ref{sec:dbs} the audio and visual features are extracted at different frame rates. Therefore they need to by synchronised. This is achieved by upsampling the visual features, to match the frame rate of the audio features (100fps), by linear interpolation similarly to \cite{Potamianos2003}.

Finally, due to randomness in initialisation, every time a deep network is trained the results are slightly different. In order to present a more objective evaluation we run each experiment 10 times and we report the mean and standard deviation of the performance measures. 

\looseness - 1

\subsection{Training}

\subsubsection{Single Stream Training} 

\textbf{Initialisation:} First, each stream is trained independently. The encoding layers are pre-trained in a greedy layer-wise manner using Restricted Boltzmann Machines (RBMs) \cite{hinton2012practical}.  Since the input (pixels or spectrograms) is real-valued and the hidden layers are either rectified linear or linear (bottleneck layer) four Gaussian RBMs \cite{hinton2012practical}  are used. Each RBM is trained for 20 epochs with a mini-batch size of 100 and L2 regularisation coefficient of 0.0002 using contrastive divergence. The learning rate is fixed to 0.001  as suggested in \cite{hinton2012practical} when visible/hidden units are linear.

As recommended in \cite{hinton2012practical} the data should be z-normalised, i.e., the mean and standard deviation should be equal to 0 and 1, respectively, before training an RBM with linear input units. Hence, each image is z-normalised before pre-training the encoding layers. Similarly, each spectrogram frame is also z-normalised.

\textbf{End-to-End Training:} Once the encoder has been pretrained then the BLSTM is added on top and its weights are initialised using glorot initialisation \cite{glorot2010understanding}.
The Adam training algorithm \cite{kingma2014adam} is used for end-to-end training with a mini-batch size of 10 utterances. The default learning rate of 0.001 led to unstable training so it was reduced to 0.0003. Early stopping with a delay of 5 epochs was also used in order to avoid overfitting and gradient clipping was applied to the LSTM layers. 

\subsubsection{Audiovisual Training} 

\textbf{Initialisation:} Once the single streams have been trained then they are used for initialising the corresponding streams in the multi-stream architecture. Then another BLSTM is added on top of all streams in order to fuse the single stream outputs. Its weights are initialised using glorot initialisation.

\noindent
\textbf{End-to-End Training:} Finally, the entire network is trained jointly using Adam with a mini-batch size of 10 utterances. Since the individual streams are already initialised at good values a lower learning rate is used, 0.0001, to fine tune the entire network. Early stopping and gradient clipping were also applied similarly to single stream training.


\section{Experiments}

In this section we report the results on OuluVS2 and AVIC databases. We have experimented with using the end-to-end audiovisual system shown in Fig. \ref{fig:system} but also with the individual streams, i.e., audio- and video-only classification. In the latter case, we just use the corresponding single stream, encoder + BLSTM. 

\subsection{Results on AVIC database}

Results for the AVIC database are shown in Table \ref{tab:resultsAVIC}. Since this is an imbalanced dataset, see section \ref{sec:dbs}, using just the classification rate can be misleading. Hence, we also report the unweighted average recall (UAR) rate and the mean F1 measure over all 4 classes. First of all, we see that the proposed end-to-end system significantly outperforms the current state-of-the-art on the AVIC database, which is based on handcrafted features and prediction-based audiovisual fusion \cite{petridis2016Pred}. It results in a statistically significant absolute mean F1 improvement of 19\% and 9.8\% for the audio-only and audiovisual classification, respectively. 

It is also clear that the audio-only classifier performs much better than the video-only classifier. This is expected since most of the information is carried by the audio channel. In addition, some vocalisations can be accompanied by subtle facial expressions, like hesitation in Fig. \ref{fig:hesitationExample_AVIC}, or even no facial expression at all. However, the visual modality is still useful and the audiovisual combination using the end-to-end model results in a statistically significant absolute improvement of 2\% of the mean F1 over the audio-only model.

\begin{table}[t]
\renewcommand{\arraystretch}{1.3}
\renewcommand{\tabcolsep}{7pt}
\caption{ Mean (standard deviation) F1, Unweighted Average Recall (UAR) and Classification Rates (CR) over 10 runs for the Audio-only classifier (A), Video-only classifier (V) and audiovisual classifiers (A + V) on the AVIC database. Subjects 1 to 7 are used as test set. The highest value in each column is shown in bold.}
\label{tab:resultsAVIC}
\centering
\begin{tabular}{cccc}
\toprule  Stream & Mean F1  & UAR & CR   \\
\midrule \multicolumn{4}{c}{Current State-of-the-art \cite{petridis2016Pred}} \\
\midrule A &  54.1 (2.2)  & 58.7 (2.4) & 58.8 (2.4) \\
V &  44.0 (2.0)  & 48.9 (2.5)   & 48.5 (2.6)  \\
 A + V  & 65.3 (2.9)  & 64.9 (3.0) & 72.6 (3.0)  \\
 \midrule \multicolumn{4}{c}{End-to-End Model} \\
\midrule A  &  73.1 (2.3)  &  72.6 (3.3)  & 79.6 (1.7)     \\
V &45.4 (5.2) & 48.4 (4.1) & 66.9 (1.4) \\
\ A + V  & \textbf{75.1 (1.5)} & \textbf{73.8 (1.5)} & \textbf{80.3 (1.5)}   \\
\bottomrule

\end{tabular} 

\end{table}

\begin{table}[t]
\renewcommand{\arraystretch}{1.3}
\renewcommand{\tabcolsep}{7pt}
\caption{Mean (standard deviation) classification rate over 10 runs of the different views and their combinations with audio on the OuluVS2 database. }
\label{tab:resultsOuluClean}
\centering
\begin{tabular}{ccc}
\toprule  Stream & V & A + V   \\
\midrule
Frontal &  91.8 (1.1) & 98.6 (0.5)   \\
30\degree & 87.3 (1.6) & 98.7 (0.5)   \\
45\degree  &  88.8 (1.4) & 98.3 (0.4)  \\
60\degree &86.4 (0.6)& 98.6 (0.6)  \\
Profile  & 91.2 (1.3) & 98.9 (0.5)  \\
\midrule
Clean Audio &  \multicolumn{2}{c}{98.5 (0.6)} \\
\bottomrule

\end{tabular} 

\end{table}

\subsection{Results on OuluVS2 database}

We consider a single view scenario where we train and test models on data recorded from a single view. Results are shown in Table \ref{tab:resultsOuluClean}.  This dataset is balanced so we just report the classification rate which is the default performance measure for this database \cite{saitoh2016concatenated}. The best performance in video-only experiments is achieved by the frontal and profile views followed by the 45\degree, 30\degree and 60\degree views. The audio-only model achieves a very high classification accuracy, 98.5\%. This is due to the audio signal being clean, without any background noise, and the participants uttering phrases which are much longer than the vocalisations on AVIC database. We also notice that audiovisual fusion does not lead to an improvement over the audio-only model. This is not surprising, given the very high accuracy already achieved by the audio classifier in clean conditions.

In order to test the benefits of audiovisual fusion we have run experiments under varying noise levels. The audio signal is corrupted by additive babble noise from the NOISEX database \cite{varga1993assessment} so as the signal-to-noise ratio (SNR) varies from 0dB to 20dB. Results are shown in Table \ref{tab:resultsOuluNoisy}. As expected, the audio model is significantly affected by the addition of noise and its performance is degraded more and more as the noise level increases leading to a classification rate of 28.4\% at 0dB. All audiovisual models significantly outperform the audio only model due to presence of the visual modality which is not affected by acoustic noise. 

It is worth pointing out, that although there are significant differences in performance between the views in the video-only case, they all result in almost the same performance 
in the audiovisual case when audio is clean, i.e. no noise added and 15/20dB. However, as the acoustic noise level increases their differences become more evident. It is interesting that the combination of noisy audio with different views does not follow exactly the same pattern as observed in Table \ref{tab:resultsOuluClean}. Between 0dB and 10dB, the combinations of audio with the 45\degree and frontal views are the best ones. The combination of audio and the 60\degree view is the worst one, which is consistent with Table \ref{tab:resultsOuluClean} but surprisingly the combination of audio and profile view also performs poorly at 0dB. This is an indication that there could be a non-linear interaction between audio and different views when the noisy levels increase but this deserves further investigation.   

We should also mention, that beyond 10\,dB the performance of the audiovisual fusion model is worse than the performance of the video-only system, which varies between 86.4\% (60\degree view) and 91.8\% (frontal view). This is probably due to the fact that the audiovisual system is trained with clean audio data. Given the very high classification accuracy achieved by the audio-only model under clean conditions, the fusion model is probably heavily biased towards audio. The fact that audiovisual fusion results in the same performance as audio-only classification under clean conditions is also an indication towards that direction. As a consequence, when the levels of acoustic noise increase the performance becomes worse than the video-only model, however it is still able to extract some useful information from the visual modality and significantly outperform the audio-only classifier.

Finally, we should also mention that we experimented with CNNs for the encoding layers but this led to worse performance than the proposed system. Chung and Zisserman  \cite{chung2016lip} report that it was not possible to train a CNN on OuluVS2 without the use of external data. Similarly, Saitoh et al. \cite{saitoh2016concatenated} report that they were able to train CNNs on OuluVS2 only after data augmentation was used. This is likely due to the rather small training set. We also experimented with data augmentation which improved the performance but did not exceed the performance of the proposed system.

\begin{table*}[t]
\renewcommand{\arraystretch}{1.3}
\renewcommand{\tabcolsep}{7pt}
\caption{Mean (standard deviation) classification rate over 10 runs on the OuluVS2 database for different noise levels. }
\label{tab:resultsOuluNoisy}
\centering
\begin{tabular}{cccccc}
\toprule  & 20dB & 15dB & 10dB & 5dB & 0dB   \\
\midrule
Audio &  96.5 (1.5) & 91.1 (3.2) & 73.3 (5.5) & 48.1 (6.6) & 28.4 (4.7)   \\
Audio + 0\degree & 97.8 (0.6) & 94.6 (0.9) & 85.1 (1.4) & 71.0 (1.3) & 57.5 (1.2) \\
Audio + 30\degree & 97.9 (0.4) & 94.2 (0.7) & 84.2 (1.4) & 70.6 (1.1) & 56.8 (3.1)   \\
Audio + 45\degree  &  97.4 (0.7) & 94.5 (1.2) & 85.8 (1.9) & 72.1 (2.6) & 58.1 (3.3)  \\
Audio + 60\degree &97.6 (0.6)& 94.2 (1.2) & 84.1 (1.3) & 69.3 (1.7) & 53.3 (3.6)  \\
Audio + 90\degree  & 97.6 (0.8) & 95.3 (1.1) & 84.8 (1.9) & 70.8 (3.2) & 53.9 (4.5)  \\

\bottomrule

\end{tabular} 

\end{table*}

\section{Conclusions}

In this work, we present an end-to-end visual audiovisual fusion system which 
jointly learns to extract features directly from the pixels and spectrograms and perform classification using BLSTM networks. Results on audiovisual classification of nonlinguistic vocalisations demonstrate that the proposed model achieves state-of-the-art performance on the AVIC database. In addition, audiovisual speech recognition experiments using different lip views on OuluVS2 demonstrate that the proposed end-to-end model outperforms the audio-only classifier for high level of acoustic noise. The model can be easily extended to multiple streams so we are planning to perform audiovisual multi-view speech recognition and investigate the influence of audio to the different views.

\section{Acknowledgements}
This work has been funded by the European Community
Horizon 2020 under grant agreement no. 645094 (SEWA)
and no. 688835 (DE- ENIGMA).

\bibliographystyle{IEEEtran}

\bibliography{mybib}

\end{document}